\documentclass[letterpaper, 10 pt, conference]{ieeeconf}  % Comment this line out if you need a4paper

\IEEEoverridecommandlockouts                              % This command is only needed if 
\usepackage{graphics} % for pdf, bitmapped graphics files
\usepackage{epsfig} % for postscript graphics files
\usepackage{mathptmx} % assumes new font selection scheme installed
\usepackage{times} % assumes new font selection scheme installed
\usepackage{amsmath} % assumes amsmath package installed
\usepackage{amssymb}  % assumes amsmath package installed
\usepackage{hyperref}
\usepackage{cite}
\usepackage{booktabs}

\title{\LARGE \bf
Achieving Precise and Reliable Locomotion with Differentiable Simulation-Based System Identification}

\author{Vyacheslav Kovalev$^{1}$, Ekaterina Chaikovskaia$^{1}$, Egor Davydenko$^{1}$, and Roman Gorbachev$^{1}$% <-this % stops a space
\thanks{$^{1}$Moscow Institute of Physics and Technology, Russia.
{\tt\small kovalev.vv@phystech.edu}}
}

\begin{document}

\maketitle
\thispagestyle{empty}
\pagestyle{empty}

%%%%%%%%%%%%%%%%%%%%%%%%%%%%%%%%%%%%%%%%%%%%%%%%%%%%%%%%%%%%%%%%%%%%%%%%%%%%%%%%
\begin{abstract}
        Accurate system identification is crucial for reducing trajectory drift in bipedal locomotion, particularly in reinforcement learning and model-based control.
        In this paper, we present a novel control framework that integrates system identification into the reinforcement learning training loop using differentiable simulation.
        Unlike traditional approaches that rely on direct torque measurements, our method estimates system parameters using only trajectory data (positions, velocities) and control inputs.
        We leverage the differentiable simulator MuJoCo-XLA to optimize system parameters, ensuring that simulated robot behavior closely aligns with real-world motion.
        This framework enables scalable and flexible parameter optimization.
        It supports fundamental physical properties such as mass and inertia. Additionally, it handles complex system nonlinear behaviors, including advanced friction models, through neural network approximations.
        Experimental results show that our framework significantly improves trajectory following.
        It reduces rotational deviation by 75\% and increases travel distance in the commanded direction by 46\% compared to a baseline reinforcement learning method.
\end{abstract}

\section*{Supplementary Material}
\noindent \textbf{Code:} \texttt{\url{https://wavegit.mipt.ru/Slavoch/mjx_sysid}}

%%%%%%%%%%%%%%%%%%%%%%%%%%%%%%%%%%%%%%%%%%%%%%%%%%%%%%%%%%%%%%%%%%%%%%%%%%%%%%%%

\section{Introduction}
Accurate simulation is crucial for the success of reinforcement learning (RL) and optimization-based control of robots \cite{di2018dynamic, kim2019highly, gaertner2021collision, hwangbo2019learning, li2024reinforcement}.
The performance of these control strategies is closely related to the fidelity of the simulation models.
A primary obstacle in achieving this fidelity is system identification, the precise estimation of a robot's physical parameters to bridge the gap between simulated and real-world behavior \cite{gajek2018mathematical}.

Traditional system identification methods rely on direct torque measurements to infer model parameters, but these approaches present significant limitations.
Torque sensors are often expensive, prone to noise, and not always available in real-world robotic setups.
Moreover, conventional identification techniques typically require controlled experimental environments, making them impractical for many robotic applications.

An alternative approach in RL is domain randomization.
Domain randomization involves training a policy across different environments with varying parameters, such as mass and control gains.
The core idea is to prevent the policy from overfitting to a single simulation environment.
However, a key disadvantage of this method is that it still requires tuning and often results in more conservative policies, sacrificing agility in the process.

\subsection{Our Approach}

\begin{figure}[t]
        \centering
        \includegraphics[width=3in]{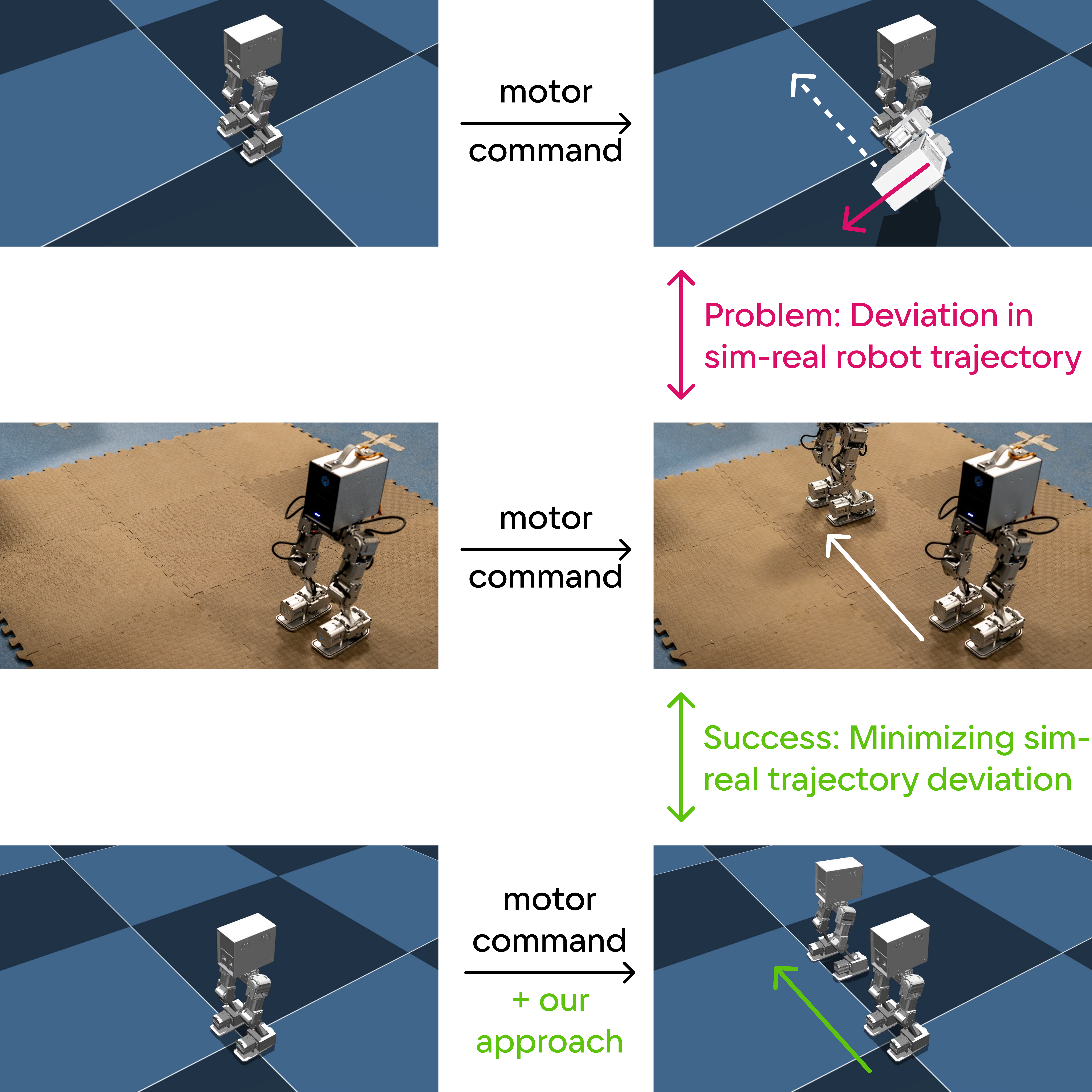} % Replace "myfigure.png" with your actual file name
        \caption{
                \textbf{Results of system identification step:}
                (Top) Simulated robot with default parameters, gradually falling over time.
                (Middle) The real robot successfully walks with natural motion.
                (Bottom) Simulated robot with optimized motor parameters, walking successfully and closely mimicking the real robot’s movement.
            }
        \label{fig:overview}
\end{figure}

% \begin{figure}[t]
%         \centering
%         %       \includegraphics[options]{name}
%         \includegraphics[width=3in]{assets/test2.pdf} % Replace "myfigure.png" with your actual file name
%         %\includegraphics[scale=1.0]{figurefile}
%         \caption{
%                 \textbf{Proposed approach for aligning simulated and real motor behavior:}
%                 (Top) Real motor in motion, demonstrating actual performance.
%                 (Bottom) Simulated motor with optimized parameters, also in motion but reaching a different angle.
%                 The optimization process addresses the deviation between the real and simulated trajectories, improving alignment.
%             }
%         \label{fig:approach}
% \end{figure}

In this work, we introduce a novel \textbf{control framework that integrates a new system identification step into the RL training loop} (as shown in \autoref{fig:overview}).
This identification step eliminates the need for torque measurements, instead estimating system parameters using only observable trajectory data (positions, velocities) and control inputs.
By leveraging the differentiable simulator MuJoCo-XLA (MJX), we optimize system parameters to minimize the discrepancy between simulated and real trajectories, creating a more accurate and adaptable simulation environment for reinforcement learning and model-based control.

\subsection{Contribution}
\begin{itemize} 
        \item \textbf{A novel control framework with a new system identification step for RL training} -- A key advantage of our framework in new \textbf{system identification step} featuring the following key properties:        
        \begin{itemize} 
                \item \textbf{No torque sensors required} -- Unlike traditional methods, our approach estimates system parameters using only trajectory data (positions, velocities) and control inputs, eliminating the need for direct torque measurements.
                \item \textbf{Optimization with a differentiable simulator} -- We leverage the differentiable simulator MJX to fine-tune system parameters, ensuring that the simulated robot behavior closely matches real-world motion.
                \item \textbf{Flexible across robotic systems} -- Since MJX supports all MuJoCo models, our method can be utilized across diverse robotic systems, regardless of their complexity.
                \item \textbf{A flexible parameter optimization} -- Enables optimization of both fundamental model parameters (e.g., mass, inertia) and more complex representations, such as neural network approximations for friction, delays, and other nonlinear system behaviors.
        \end{itemize} 
        \item \textbf{Experimental validation of our system identification approach} -- We validate our approach by collecting real motor trajectory data and show that optimizing motor parameters using our method significantly improves the alignment between simulated and real motor trajectories.
        \item \textbf{Compared simulation-to-reality alignment and RL performance} -- The identified motor parameters were used in policy learning and compared to the standard RL approach.
        Results demonstrate that our framework reduced rotational deviation by 75\% and improved travel distance in the commanded direction by 46\%.
\end{itemize}

\section{Related Work}
\label{sec:related_work}

System identification plays a crucial role in building accurate robot simulations and controllers, particularly in RL and optimization-based control frameworks.
Traditional system identification methods, especially for electric motors, face significant challenges due to the complexities of motor dynamics, making it difficult to develop precise models for real-world applications.

\subsection{Traditional system identification methods}
System identification has long relied on direct torque measurements to model motor dynamics accurately.
These measurements are typically obtained from external sensors or dedicated test stands that provide precise data under controlled conditions \cite{gajek2018mathematical, huang2022review, urrea2021dynamic,mamedov2020practical}.
While these setups enable accurate parameter estimation for actuator models, they come with several drawbacks.
First, they require specialized hardware, which increases costs and is prone to noise and calibration errors.
Second, real-time measurements during robot operation are difficult due to sensor limitations and dynamic environmental factors.

Motor modeling is further complicated by inherent nonlinearities, friction, and transmission effects that impact motor performance \cite{gajek2018mathematical, qin2024dynamics}.
Several methods address these challenges, ranging from analytical least-squares-based parameter identification \cite{huang2022review} to more data-driven approaches such as neural network-based approximations of motor dynamics \cite{hwangbo2019learning}.
ActuatorNet, for instance, learns to approximate full motor dynamics during robot operation using torque data\cite{rudin2022learning}.

However, the dependence on torque measurements remains a significant limitation in practical robotics applications.

\subsection{Addressing the sim-to-real gap in RL}
In reinforcement learning, the accuracy of the underlying system model directly affects policy training efficiency and performance \cite{hwangbo2019learning, xie2023learning}.
One popular strategy to address the sim-to-real gap is domain randomization, where key simulation parameters are randomized during training to improve policy robustness to real-world variations \cite{akkaya2019solving, li2024reinforcement}.
Although effective in generating robust policies, this approach does not enhance the simulation model itself and often results in overly conservative behaviors \cite{tiboni2023domain}.

Transfer learning in RL involves leveraging pre-trained policies from one environment or task and adapting them to a new setting, such as fine-tuning a policy trained in simulation for real-world deployment \cite{siekmann2021sim}.
While this reduces the real-world data requirements, it still relies on accurate simulations for initial training.
Offline RL methods further attempt to overcome these challenges by learning policies from pre-collected datasets without interacting with the real system~\cite{wu2023daydreamer}.
However, the performance of offline RL is highly dependent on the quality and diversity of the dataset.

Despite these advances, many RL applications still rely on simplified system models, leading to performance degradation when deployed on real robots.
Addressing these limitations requires better identification techniques for system dynamics \cite{xie2023learning}.
\subsection{Differentiable simulators}

Differentiable simulators such as Brax, TDS, and MJX enable gradient-based system identification \cite{brax2021github, heiden2021neuralsim}.
Among them, MJX, built on MuJoCo, is the most widely used RL simulator \cite{kaup2024review}, according to 2023 statistics.

Comparative studies \cite{liao2023performance, erez2015simulation} have shown that MuJoCo outperforms other simulators in terms of accuracy and numerical stability across various experiments.
Additionally, MuJoCo offers advanced physics modeling, including unique features such as tendon simulation, which are not commonly found in other simulators or frameworks.

Its widespread adoption in robotics research \cite{xie2018feedback, xie2019iterative, li2024reinforcement} and strong community support further enhance its reliability and continuous development.

Given these advantages, we chose MuJoCo as the foundation for our work.
\subsection{Our contribution in the context}
In summary, the feasibility of system identification using differentiable simulators was demonstrated by TDS.
Building on this foundation, our approach introduces a novel, scalable, and torque-measurement-free system identification method that leverages differentiable simulation, seamlessly integrated into the RL training loop.

By leveraging the advantages of MJX, we offer a robust and practical solution that is validated through real-robot experiments, ensuring its effectiveness in real-world applications.
\section{Problem Formulation}
\begin{figure}[t]
        \centering
        \includegraphics[width=3in]{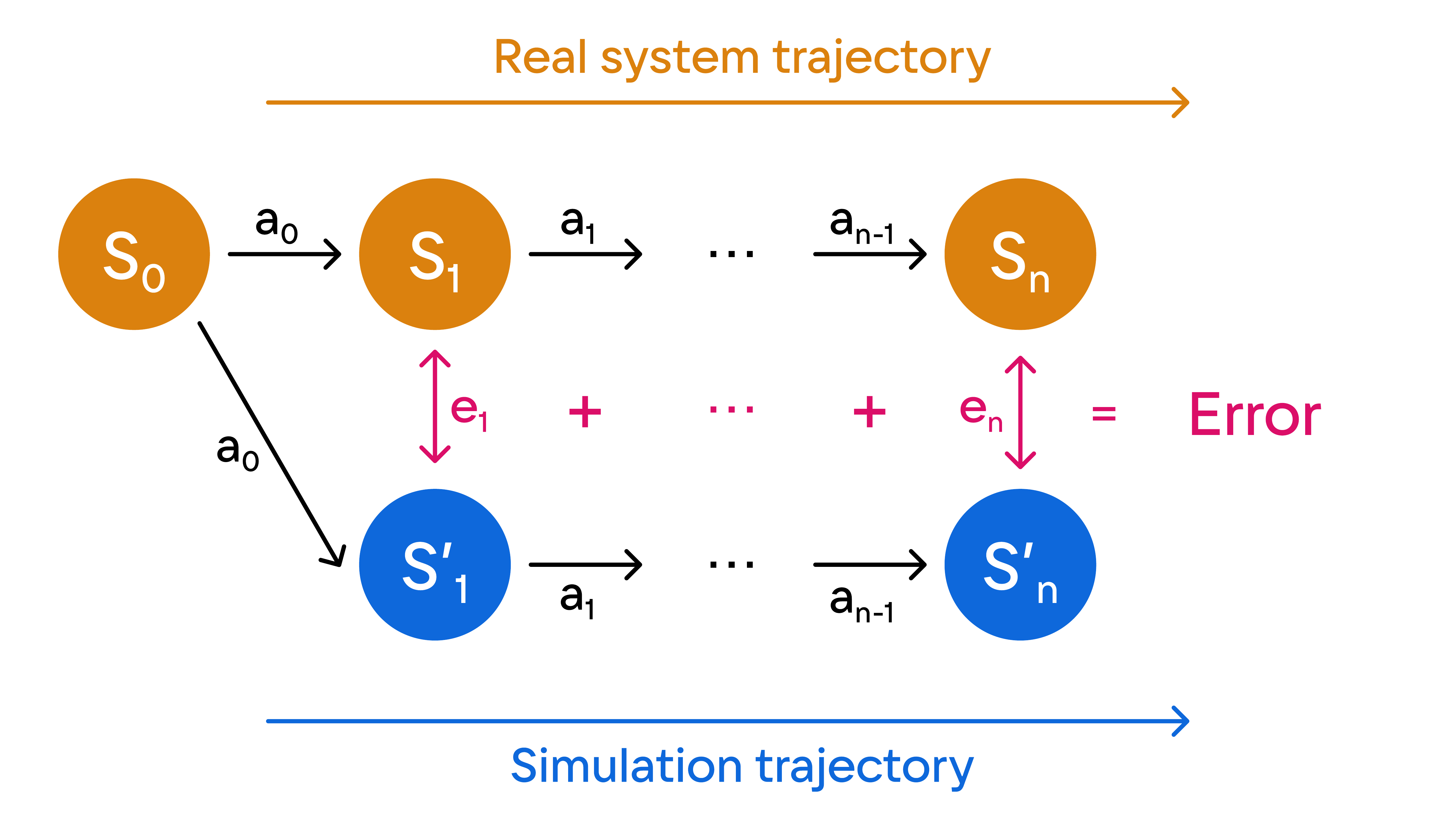} % Replace "myfigure.png" with your actual file name
        \caption{\textbf{Illustration of the main optimization problem:}
        The real system follows a trajectory represented by the sequence of states $\{s_i\}$, with each transition governed by the action input $a_i$.
        By simulating the robot's behavior, we generate a corresponding simulated trajectory, where $s'_i$ represents the simulated counterpart of the real state $s_i$.
        At each step, the error $e_i$ is the difference between the real and simulated states.
        The total discrepancy between the real and simulated trajectories is given by the sum of these errors, which the optimization process (\autoref{eq:main_opt_problem}) seeks to minimize.}
        \label{fig:problem_statement}
\end{figure}

        In this section, we formally define the problem to be solved.

        Let $s_i$ and $a_i$ denote the system's state and control input, respectively, at timestep i.
        We assume that there exists a unique parameter set $z^*$ (such as inertia) capable of perfectly modeling the system's dynamics, resulting in a predicted next state $s'_{i+1}$ that is identical to the actual next state $s_{i+1}$.
        \begin{equation}
                s_{i+1} = s'_{i+1} = \Phi_{z^*}(s_i, a_i,\delta),
                \label{eq:next_state}
        \end{equation}
        where $\delta$ is the prediction step size, and $\Phi_z(s, a,\delta)$ represents the integration scheme.
        $\delta$ should be chosen small enough to ensure the desired level of accuracy while avoiding excessively small values that would increase computational cost \cite{aeran2016time}.
        \autoref{eq:next_state} implies that the next state predicted by the simulation should match the real next state if the model is accurate.

        Thus the optimization problem is formulated as:
        \begin{equation}
                \begin{aligned}
                        z^* = \arg &\min_z \sum_{i=1}^N ||s'_i - s_i||_2^2 \quad \text{s.t.} \\
                        s'_{i+1} &= \Phi_z(s'_i, a_i,\delta), \\
                        s'_0 &= s_0.
                \end{aligned} 
                \label{eq:main_opt_problem}
        \end{equation}
        The objective is to minimize the difference between the simulated and real state trajectories, thereby identifying the optimal system parameters.
        A visual description of this process is shown in \autoref{fig:problem_statement}.

        While this formulation effectively identifies system parameters, numerical integration errors may accumulate over time, degrading the optimization process.
        To mitigate this, we divide the trajectory into multiple segments of length $N$ and minimize the error within each segment:
        \begin{equation}
                \begin{aligned}
                        z^* = \arg &\min_z \sum_{j=0}^M \sum_{i=1}^N ||s'_{i,j} - s_{i,j}||_2^2 \quad \text{s.t.} \\
                        s'_{i+1,j} &= \Phi_z(s'_{i,j},a_{i,j},\delta), \\
                        s'_{0,j} &= s_{0,j}, 
                \end{aligned} 
                \label{eq:final_opt_problem} 
        \end{equation}
        where $j$ is the segment index, $i$ is the time step within each segment, and $M$ is the total number of segments.
        By selecting an appropriate $\delta$ (prediction step size) and $N$ (prediction horizon for each segment), we aim to minimize the impact of integrator error on the optimization process.
        
        \subsection{Limitations}
        This approach assumes that the primary difference between simulation and reality is captured by the parameter vector $z$.
        However, if unmodeled factors - such as variations in mass, inertia, or external disturbances - are present and not encoded in $z$, then \autoref{eq:next_state} may no longer hold, leading to inaccuracies in state predictions.

        Furthermore, due to the inherent complexity of dynamics modeling, achieving the exact equality defined in \autoref{eq:next_state} may be impractical.
        However, our proposed method remains effective as long as $\Phi_{z^*}(s'_i, a_i,\delta)$ provides a sufficiently close approximation of the true system dynamics.
\section{Methodology}
\label{sec:methodology}

In this section, we detail the practical implementation of the optimization problem defined in \autoref{eq:final_opt_problem}.

To enhance the robustness of RL for robot locomotion, we focus on identifying motor dynamics.
Our experiments are based on the Mini $\pi$ robot~\cite{high2025torque} and its motors.

From our experiments, we found that a neural network parameterized by $z$ could approximate motor output torque.
However, this approach has a significant drawback - high variance in predictions, especially in data-sparse regions.
Collecting a sufficiently diverse dataset requires operating the motor in various states, which can be inherently risky and potentially damaging.

An alternative approach is to identify built-in motor parameters.
While this method lacks the flexibility of neural networks, it does not rely on an extensive dataset.
Given these trade-offs, we chose to optimize MuJoCo's built-in motor parameters $z$, specifically \textbf{armature}, \textbf{frictionloss}, and \textbf{damping}.

In the following sections, we detail the methodology for motor dataset collection, describe the parameter optimization process, present a comparative analysis of simulation reliability, and demonstrate the training of a RL policy using the optimized model.

\subsection{Dataset generation}
\label{sec:dataset_gen}
\begin{figure}[t]
        \centering
        \vspace{0.1cm} 
        \includegraphics[width=3in]{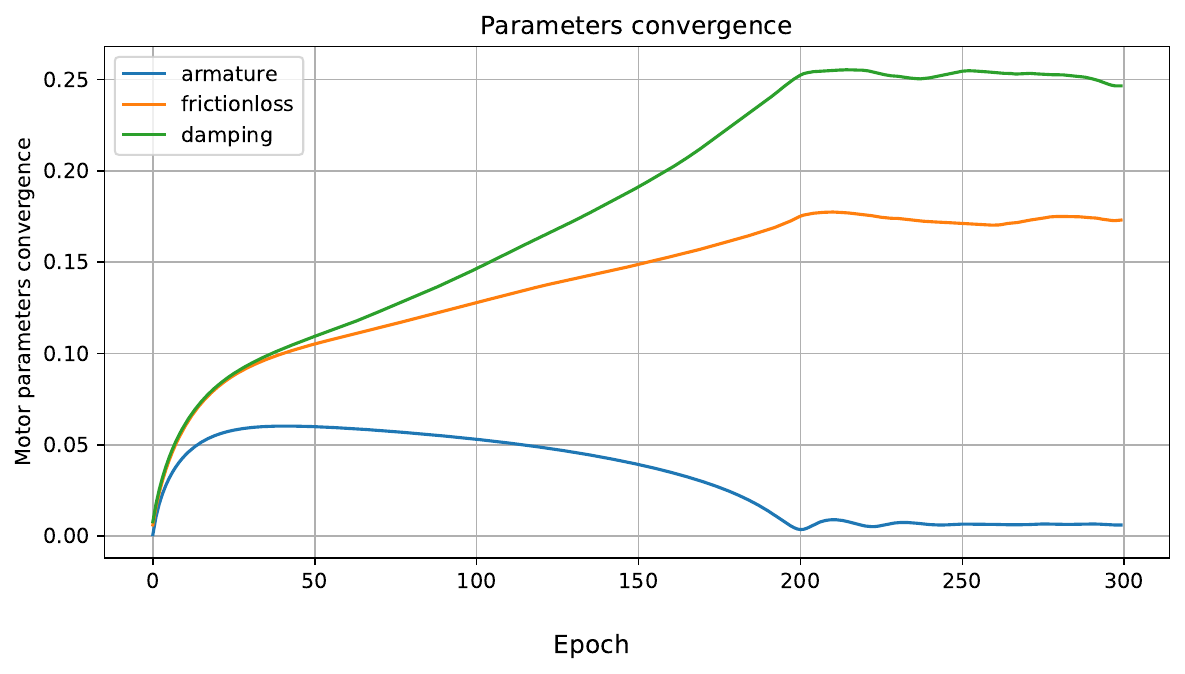}
        \caption{
                \textbf{Parameters convergence:}
                Motor model parameters (frictionloss, damping, armature) converge smoothly over epochs.}
        \label{fig:loss}
\end{figure}
First, we extracted a real motor from the Mini $\pi$ (HTDW-5047-36-NE~\cite{high2025torque}).
We then securely mounted the setup to a table.
As the motor's load, we used a rod with a known mass and inertia.

The optimization problem in \autoref{eq:final_opt_problem} requires real-world trajectories of the system.
To generate these, we create a sequence of desired velocities $\{v_i^{des}\}$ using a sum of Fourier modes.
The number of modes, their coefficients, and frequencies are randomly sampled within a predefined range to ensure a diverse set of velocity trajectories.
The generated velocities are then clipped to the motor's maximum velocity (a hardware-specific constraint).
Finally, we integrate these velocities to obtain the corresponding desired angles $\{q_i^{des}\}$.

The input command for our motor corresponds to the desired angles, defined as:
\begin{equation}
a_i = q^{des}_i.
\end{equation}
The input command is then processed by a PD controller.
By default, our robot successfully walks with proportional (P) gains in the range of [40-80].
We anticipate that reducing the PD gains will enhance the prominence of the motor's intrinsic dynamics.
To achieve this, we set the proportional gain to $K_p=20$ and the derivative gain to $K_d=1$.

Next, the action sequence is applied to the real motor through a PD controller, generating the actual state sequence $\{s_i\}$, where: 
\begin{equation}
s_i=[q_i,v_i]^T
\end{equation}
is the concatenation of the real motor's measured angles and velocities at timestep $i$.
Finally, we resample the entire dataset to a 1~ms time step ($\delta$ in \autoref{eq:final_opt_problem}).
\subsection{Motor identification results}

The full dataset was split into training and testing sets.
The training dataset was further divided into segments of length $N=4$ for optimization in \autoref{eq:final_opt_problem}.
For the integration function $\Phi_z(s,a,\delta)$, we use MuJoCo's default explicit Euler integrator.
While higher-order methods like RK4 offer better numerical stability, we found that Euler integration results in four times faster convergence with no significant loss in accuracy.

Optimal motor parameters were obtained by minimizing the objective function (\autoref{eq:final_opt_problem}).
\autoref{fig:loss} illustrates the smooth convergence of the motor model parameters to their optimal values during the optimization process, demonstrating the stability and effectiveness of our method.
It is crucial to note that MJX is highly sensitive to initial system parameters.
Unrealistic initial guesses can lead to instability in next-state predictions.
A comparison of the loss function using the optimal parameters $z^*$ and the initial guess revealed an approximate 20\% reduction.
\begin{figure}[t]
        \centering
        \vspace{0.1cm} 
        \includegraphics[width=3in]{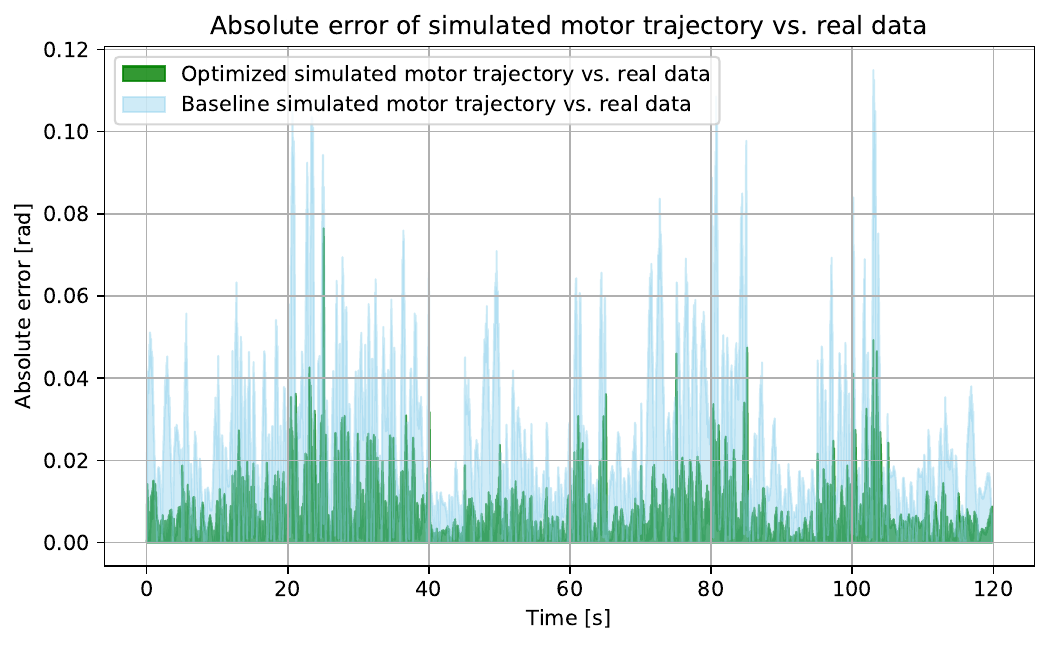}
        \caption{
                \textbf{Absolute error between simulated and real motor trajectories:}
                The optimized motor model (green) shows a reduced error compared to the baseline model (blue), demonstrating closer trajectory alignment of the optimized motor model.}
        \label{fig:motor_sim_real_qpos_error}
\end{figure}
Next, we initialized the simulated motor with the optimized parameters, matching the initial state of the real motor, and applied control inputs from a previously unseen test dataset.
The baseline model, characterized by zero frictionloss and damping, and minimal armature (to mitigate excessive oscillations), served as a comparative benchmark.
We aimed to demonstrate the significant improvement achievable with our simple identification step compared to this baseline model.

As shown in \autoref{fig:motor_sim_real_qpos_error}, the optimized model exhibited a significantly improved alignment with the real motor trajectory compared to the baseline model.
The mean squared error (MSE) between the optimized simulated motor trajectory and the real trajectory was approximately 8 times smaller than the MSE of the baseline model.

\subsection{Validation with real robot}
\begin{figure*}[htpb]
        \centering
        \includegraphics[width=\textwidth]{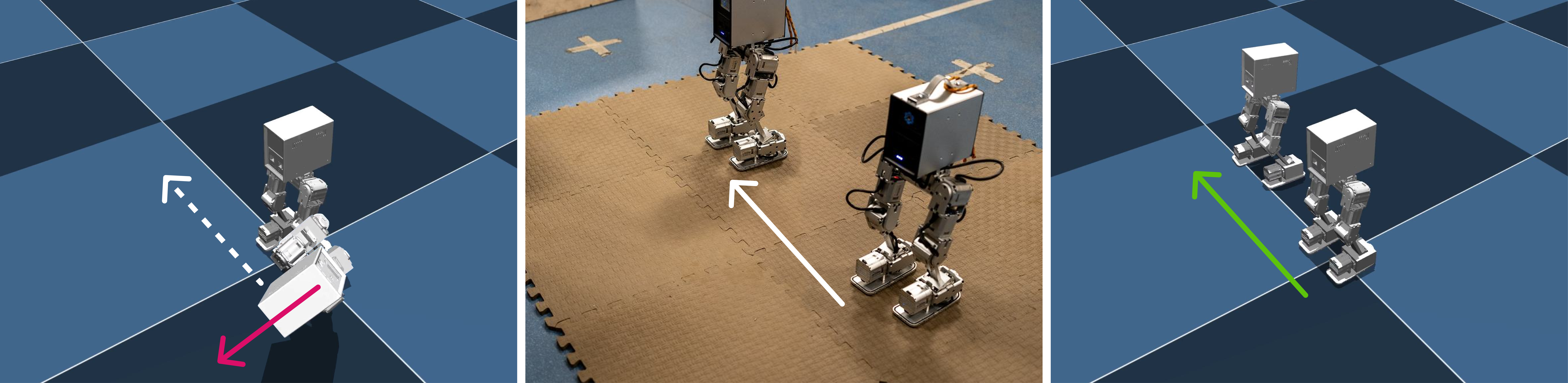}  % Full-page width
        \caption{
            \textbf{Comparison of robot trajectories:}
            The solid arrow indicates the direction of movement.
            (Left) Baseline simulated robot, which fails to replicate the real robot's movement and falls instead.
            (Middle) Real robot trajectory.
            (Right) Simulated robot with optimized motor parameters, showing close alignment with real robot behavior.
        }
        \label{fig:policy1_comparison}
    \end{figure*}

        We aim to evaluate the effectiveness of integrating learned motor parameters into a MuJoCo robot model to improve simulation accuracy.
        The baseline model assumes zero values for joint parameters, reflecting the common RL practice of domain randomization instead of precise model parameter identification.

        \textbf{Description:}
        To test our approach, we incorporated the learned motor parameters into the robot MuJoCo simulation.
        The baseline model, by contrast, used joint parameters set to zero.
        This comparison allows us to assess whether our optimized model better replicates real-world robot behavior.
        
        \textbf{Setup:}
        We employed a pre-trained RL policy and tested it in three settings: the real world, a simulation with the baseline model, and a simulation with the optimized model.
        Each experiment was repeated 10 times, with each run lasting 7 seconds.
        During these trials, the policy relied exclusively on time, joint positions, and joint velocities, with all control inputs fixed at zero.

        \textbf{Results:}
        As shown in \autoref{fig:policy1_comparison}, the real robot exhibits a gradual drift over time, which our optimized simulated model successfully replicates.
        In contrast, the baseline model fails to capture this behavior and instead falls.

        To quantify simulation accuracy, we measured the mean Euclidean distance between the robot's starting position and final position.
        Our optimized model deviates by 0.16 meters, while the baseline model deviates by 1.13 meters, as detailed in \autoref{tab:results}.
        These results demonstrate an 86\% improvement in the accuracy of the robot's displacement.

\subsection{Applying the optimized model to RL policy learning}

        We aim to demonstrate that our framework improves RL policy learning and enhances real-world robot performance.
        Specifically, we trained a robot to walk using learned motor parameters and evaluated its performance compared to a default RL approach.

        \textbf{Description:}
        To evaluate the effectiveness of our framework, we trained a walking policy using the optimized robot model and compared it to a policy trained on the baseline model.
        The baseline represents the default RL approach, where system parameters are not explicitly identified and remain unknown.
        This comparison allows us to assess the impact of incorporating learned motor parameters on policy performance.
        
        \textbf{Setup:}
        Training was conducted using the LeggedGym~\cite{rudin2022learning} framework with its built-in reward functions, without applying any randomization techniques.
        The policy was trained on the optimized robot model and compared against a baseline policy trained using the unadjusted model.

        After training, both policies were deployed on the real robot under identical conditions.
        The only modification was reducing the angular velocity scale in the real system to half of the simulation value to account for IMU noise inaccuracies.
        Additionally, we set the commanded linear velocity to its maximum.

        The experiment with each policy was repeated 10 times, with each run lasting 7 seconds.
        During these runs, we measured the following metrics: mean travel distance in the commanded direction and mean rotational deviation from the original heading.

\begin{figure}[hb]
                \centering
                \includegraphics[width=3in]{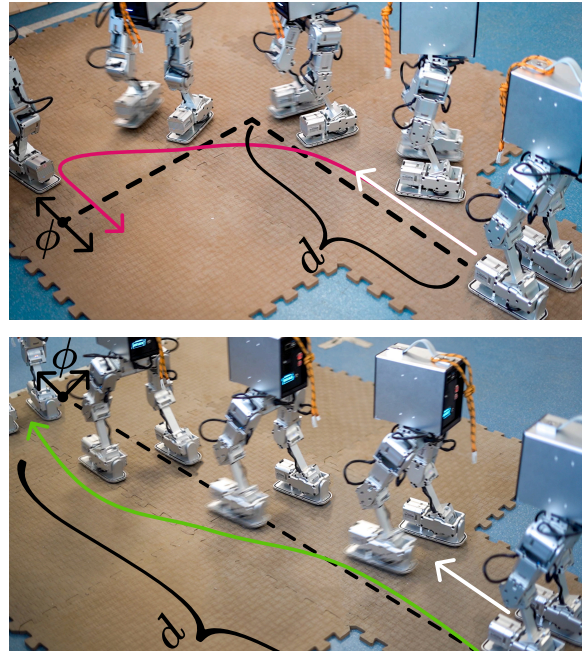} % Replace "myfigure.png" with your actual file name
                \caption{\textbf{Robot Performance Comparison: Refined vs. Baseline RL Policies.}
                The traveled distance is denoted as $d$, and the rotation deviation from the original direction is represented as $\phi$.
                The top image shows robot performance using the baseline model, exhibiting excessive rotation and reduced travel distance.
                The bottom image shows the robot's performance using the refined model, demonstrating improvements in both travel distance and rotational alignment.}
                \label{fig:rl_comparison}
        \end{figure}

        \textbf{Results:}
        As shown in \autoref{fig:rl_comparison}, the baseline policy caused excessive rotation during walking, limiting the robot's forward travel.
        The baseline policy resulted in the robot traveling only 1.12 meters in the commanded direction (\autoref{tab:results}), with a rotational deviation of 109 deg from its original orientation.

        In contrast, the policy trained on the refined motor model enabled the robot to travel 1.64 meters, with a rotational deviation of only 27 deg.
        This demonstrates that our proposed framework improved the robot's walking performance by 46\% and reduced rotational deviation by 75\%.

        These results highlight that our adjusted model leads to more stable and efficient locomotion in real-world deployment.

\section{Conclusion}
In this work, we introduced a framework for identifying and adjusting system dynamics using only state sequences (positions and velocities) and control inputs.
By leveraging the differentiability of MJX with respect to model parameters, we optimized key motor properties such as frictionloss, damping, and armature using real-world data.

Our results demonstrate that the adjusted motor model significantly improves alignment between simulated and real motor trajectories compared to a baseline model with default parameters (\autoref{tab:results}).
When integrated into a full robotic system, our method improved trajectory replication, significantly reducing the gap between simulation and real-world behavior.

Additionally, we trained an RL policy using the adjusted model and observed improved real-world performance over a baseline policy trained on an unoptimized model (\autoref{tab:results}).

\begin{table}[bpht]
        \centering
        \caption{Comparison of our method vs. baseline}
        \begin{tabular}{lccc}
                \toprule
                \textbf{Metric} & \textbf{Baseline} & \textbf{Ours} & \textbf{Improvement} \\
                \midrule
                \multicolumn{4}{c}{\textbf{Sim-to-Real Validation}} \\
                Mean position drift (m) ↓ & 1.13 & 0.16  & +86\%  \\
                \midrule
                \multicolumn{4}{c}{\textbf{RL Policy Comparison}} \\
                Mean rotation deviation (deg) ↓ & 109  & 27  & +75\%  \\
                Mean travel distance (m) ↑ & 1.12  & 1.64  & +46\%  \\
                \bottomrule
        \end{tabular}
        \label{tab:results}
\end{table}
    
\section{Discussion and Future Work}

While our framework effectively identifies simulated motor parameters, more complex system behaviors could be better approximated using neural networks.
Incorporating historical motor data could help capture unobservable system states.
This may assist in approximating temperature-dependent effects and command delays.
Additionally, it could help denoise input signals, improving overall system accuracy.

Beyond individual motor identification, this framework could be extended to learn other system parameters, making it a powerful tool for system identification and parameter tuning.
Notably, our framework is also applicable to motors with high gear ratios, where traditional methods may struggle, further expanding its utility.

Additionally, for robustness in RL applications, our approach provides a foundation for domain randomization.
Even if system parameters cannot be identified precisely due to noisy sensors, the estimated parameters could still be used within a constrained domain randomization range.
By limiting parameter variations, policies learned with this approach may be less conservative compared to standard domain randomization techniques.

\addtolength{\textheight}{-12cm}

%%%%%%%%%%%%%%%%%%%%%%%%%%%%%%%%%%%%%%%%%%%%%%%%%%%%%%%%%%%%%%%%%%%%%%%%%%%%%%%%

\bibliographystyle{IEEEtran}
\bibliography{references}

\end{document}